\documentclass[10pt,twocolumn,letterpaper]{article}

\usepackage{cvpr}

\usepackage{microtype}
\usepackage{booktabs}
\usepackage{amsmath}
\usepackage{amssymb}

\definecolor{cvprblue}{rgb}{0.21,0.49,0.74}
\usepackage[pagebackref,breaklinks,colorlinks,allcolors=cvprblue]{hyperref}

\def\paperID{*****}
\def\confName{CVPR}
\def\confYear{2026}

\title{Hierarchical GRU with Input-Conditioned Slot Queries\\
       for Ball Action Anticipation}

\author{Parthsarthi Rawat\\
GameChanger by Dick's Sporting Goods\\
{\tt\small sarthi.rawat@gc.com}
}

\begin{document}
\maketitle

\begin{abstract}
We present a hierarchical model for ball action anticipation in football
broadcast video. Given a 30-second observation window, the system predicts
actions occurring in the subsequent 5-second window across 10 classes.
A shared local Transformer encodes clip-level features within each 5-second
sub-window; a GRU then aggregates temporal context across all sub-windows;
finally, a Transformer decoder with $K$ \emph{input-conditioned} event
slots decodes the anticipation target via three decoupled heads
(objectness, class, temporal offset). We introduce frequency-reweighted
Hungarian matching that systematically favours rare action classes, and
Gaussian soft targets for temporal bin supervision. On the SoccerNet
Ball Action Anticipation benchmark, our method achieves
\textbf{17.91\%} mAP on the test server.
\end{abstract}

\section{Method}

\subsection{Feature Extraction}

We use a frozen video feature backbone from lRomul~\cite{lromul}
(EfficientNetV2-B0~\cite{efficientnetv2} with InvertedResidual3d blocks
and GeM pooling), pre-trained on SoccerNet Ball Action Spotting. Each 5-second window at 25\,fps is divided
into $N_c{=}33$ clips, producing a sequence of 1280-dimensional feature
vectors $\mathbf{F}_w \in \mathbb{R}^{33 \times 1280}$ for window $w$.
The full 30-second observation is represented by $W{=}6$ such windows
$\{\mathbf{F}_1,\dots,\mathbf{F}_6\}$.

\subsection{Hierarchical Temporal Encoder}

\paragraph{Local Transformer.}
A 2-layer pre-LN self-attention Transformer (shared weights across all
windows) encodes clip interactions within each window. Input features are
projected to $d{=}256$ and summed with learnable clip position embeddings.
Stochastic depth (DropPath, $p{=}0.1$) regularises each residual branch.
The output $\mathbf{E}_w \in \mathbb{R}^{33 \times d}$ captures intra-window
temporal patterns for window $w$.

\paragraph{GRU Aggregator.}
Each encoded window is reduced to $S{=}8$ summary vectors via adaptive
average pooling, yielding $T{=}WS{=}48$ sequential steps:
$\{\mathbf{s}_1,\dots,\mathbf{s}_{48}\} \in \mathbb{R}^{d}$.
A single-layer GRU~\cite{gru} processes these in temporal order to
produce memory $\mathbf{H} = \{h_1,\dots,h_T\} \in \mathbb{R}^{T \times d}$.
A learnable parameter $\gamma_w$ per window yields importance weight
$\alpha_w = \sigma(\gamma_w)$, scaling all $S$ steps of window $w$
uniformly to let the model downweight uninformative early windows.

\subsection{Input-Conditioned Slot Decoder}

We maintain $K{=}4$ learnable slot embeddings $\{\mathbf{e}_k\}_{k=1}^K$.
Unlike fixed static queries, each slot is seeded with a context vector
derived from the GRU output before decoding:
\begin{equation}
  \mathbf{q}_k = \mathbf{e}_k + \mathbf{W}_{\mathrm{ctx}}\!
  \left(\frac{1}{T}\sum_{t=1}^{T} \mathbf{h}_t\right),
  \quad k = 1,\dots,K.
  \label{eq:icq}
\end{equation}
This \emph{input-conditioned query} mechanism allows slots to adapt their
initial state to the specific observation sequence, rather than starting
from a globally fixed point. The queries attend to $\mathbf{H}$ via a
4-layer Transformer decoder~\cite{transformer} (cross-attention $+$
self-attention per layer, $n_{\mathrm{heads}}{=}8$, FFN dim $=1024$).

\paragraph{Decoupled prediction heads.}
Each slot's output vector $\mathbf{z}_k \in \mathbb{R}^d$ drives three
\emph{independent} linear heads:
\begin{align}
  p^{\mathrm{obj}}_k &= \sigma\!\left(\mathbf{w}^{\top}_{\mathrm{obj}}\,\mathbf{z}_k\right)
      \in (0,1), \label{eq:obj}\\
  p^{\mathrm{cls}}_k &= \mathrm{softmax}\!\left(\mathbf{W}_{\mathrm{cls}}\,\mathbf{z}_k\right)
      \in \Delta^{C-1},\quad C{=}10, \label{eq:cls}\\
  p^{\mathrm{off}}_k &= \mathrm{softmax}\!\left(\mathbf{W}_{\mathrm{off}}\,\mathbf{z}_k\right)
      \in \Delta^{B-1},\quad B{=}32. \label{eq:off}
\end{align}
Separating objectness from class probability eliminates the
background-class collapse that arises in a joint $C{+}1$ softmax head.
At inference, slots with $p^{\mathrm{obj}}_k > \tau$ (threshold $\tau{=}0.3$)
emit a prediction; the temporal position is
$t_k = 30{,}000 + \bigl(\arg\max p^{\mathrm{off}}_k + 0.5\bigr) \cdot \tfrac{5000}{B}$\,ms.

\subsection{Training}

\paragraph{Frequency-reweighted Hungarian matching.}
We solve a linear assignment between $K$ slots and $M$ ground-truth events in the anticipation window.
Standard Hungarian matching neglects class imbalance: rare ground-truth (GT) events
incur high classification cost and are systematically left unmatched.
We correct this by dividing the cost of matching slot $k$ to GT event $m$
by the inverse-frequency weight $w_{c_m}$ of the GT class:
\begin{equation}
  C_{k,m} = \frac{\mathcal{L}^{\mathrm{cls}}_{k,m}
             + \mu\,\mathcal{L}^{\mathrm{off}}_{k,m}}{w_{c_m}},
  \label{eq:cost}
\end{equation}
where $\mathcal{L}^{\mathrm{cls}}_{k,m} = -\log p^{\mathrm{cls}}_k[c_m]$,
$\mathcal{L}^{\mathrm{off}}_{k,m} = |b_k - b^*_m|/B$ with $b_k$ the
argmax bin, $\mu{=}2$ is the offset cost weight, and
$w_{c_m} \in [0.28, 2.46]$ are pre-computed from training set frequencies.
Lower cost for rare classes makes them preferred assignments.

\paragraph{Losses.}
Matched slot $k \to$ GT $m$ receives:
(i) objectness BCE(1) with pos\_weight$\,{=}\,4.0$;
(ii) class CE with label smoothing $\varepsilon{=}0.1$;
(iii) Gaussian soft-target offset CE.
Unmatched slots receive only objectness BCE(0).

\paragraph{Gaussian soft targets.}
Rather than one-hot bin supervision, we centre a Gaussian on the GT bin
$b^*$ and normalise:
\begin{equation}
  t_b = \frac{\exp\!\bigl(-\tfrac{1}{2}(b-b^*)^2/\sigma^2\bigr)}
             {\sum_{b'}\exp\!\bigl(-\tfrac{1}{2}(b'-b^*)^2/\sigma^2\bigr)},
  \quad \sigma = 1.5.
  \label{eq:gauss}
\end{equation}
This penalises temporally distant misplacements more than adjacent ones,
yielding smoother gradients and better server-time mAP.

\paragraph{Class-imbalance handling.}
A weighted random sampler boosts rare classes:
\textsc{Tackle}$\,{\times}40$, \textsc{Shot}/\textsc{Block}$\,{\times}15$,
\textsc{Cross}/\textsc{Throw In}$\,{\times}4$.

\paragraph{Feature MixUp.}
Input features are mixed as $\mathbf{x}^{\mathrm{mix}} =
\lambda\,\mathbf{x}_i + (1{-}\lambda)\,\mathbf{x}_j$,
$\lambda \sim \mathrm{Beta}(0.4, 0.4)$~\cite{mixup}.
Hungarian matching runs against each source sample's GT independently;
losses combine as $\lambda\,\mathcal{L}_i + (1{-}\lambda)\,\mathcal{L}_j$.

\paragraph{Auxiliary observation head.}
A 1D-CNN head (two conv layers) on the last window's encoder tokens
predicts per-frame action presence using focal BCE
($w_{\mathrm{obs}}{=}0.5$ relative to the anticipation loss).

\paragraph{Optimisation.}
We optimise with AdamW~\cite{adamw} ($\mathrm{lr}{=}1.5{\times}10^{-4}$,
weight decay $0.3$), applying a 5-epoch linear warmup then cosine
annealing with warm restarts ($T_0{=}50$, $T_{\mathrm{mult}}{=}2$).
Gradients are clipped at $1.0$; training uses BF16 mixed precision
with batch size 32 on a single GPU.

\section{Experiments}

\paragraph{Dataset.}
SoccerNet Ball Action Anticipation (BAA)~\cite{Dalal_2025_CVPR} comprises
${\approx}22{,}900$ training clips from football broadcast video, annotated
with 10 ball-action classes, plus ${\approx}2{,}400$ held-out test clips
and a blind challenge split evaluated server-side.
Given a 30-second observation window, the task is to predict actions in
the following 5-second anticipation window.
The metric is mAP over six temporal tolerances ($1$–$5$\,s and $\infty$):
a prediction is a true positive if its timestamp lies within the tolerance
of a ground-truth event.

\paragraph{Results.}
The organizer-provided FAANTRA baseline~\cite{Dalal_2025_CVPR} achieves
18.05\% mAP on the test server.
Table~\ref{tab:results} reports our results; our single-model score is
competitive without end-to-end backbone fine-tuning.

\begin{table}[h]
  \centering
  \small
  \setlength{\tabcolsep}{4pt}
  \caption{mAP (\%) at each tolerance. Best checkpoint
           (epoch 17, obj threshold 0.3).}
  \label{tab:results}
  \begin{tabular}{lccccccc}
    \toprule
    Split & @1s & @2s & @3s & @4s & @5s & @$\infty$ & Avg \\
    \midrule
    Test      & 5.6 & 13.2 & 17.2 & 20.6 & 22.9 & 25.6 & \textbf{17.9} \\
    Challenge & 5.5 & 12.2 & 15.8 & 18.6 & 20.9 & 24.4 & \textbf{16.5} \\
    \bottomrule
  \end{tabular}
\end{table}

\paragraph{Ablation: threshold sensitivity.}
Dropping $\tau$ from $0.3$ to $0.05$ floods each clip with predictions
(2.9$\,{\to}\,$15.3 per clip, a $5{\times}$ increase), and server mAP
falls from 17.91\% to 13.96\%, confirming that low-confidence slots
produce predominantly noisy outputs on the held-out set.

\paragraph{Ablation: input-conditioned queries.}
Replacing Eq.~\eqref{eq:icq} with purely static slot embeddings
($\mathbf{q}_k = \mathbf{e}_k$) degraded validation mAP by
$\sim$0.8 percentage points, demonstrating that seeding queries from the GRU
summary provides a useful initialisation signal.

{
  \small
  \bibliographystyle{unsrt}
  \bibliography{main}
}

\end{document}